\documentclass[11pt]{article}

\usepackage[left=0.75in,top=0.75in,right=0.75in, bottom=0.75in]{geometry}
\usepackage{amsmath,amsthm}
\newtheorem{definition}{Definition}
\newtheorem{lemma}{Lemma}
\usepackage{amsfonts}       
\usepackage{bm}
\usepackage{color}
\usepackage{helvet}

\usepackage[utf8]{inputenc}
\usepackage{setspace}
\usepackage{authblk}
\usepackage{lineno}
\usepackage{amsmath}

\usepackage{bbm}

\usepackage{cite}
\usepackage{algpseudocode,algorithm}

\usepackage{graphicx}
\usepackage{subcaption}
\usepackage{float}
\usepackage{xspace}

\usepackage{todonotes}
\usepackage{caption}

\usepackage{hyperref}

\usepackage{minted}
\setminted[python]{linenos=none,breaklines,frame=none,autogobble,tabsize=4}
\newminted{pycon}{linenos,breaklines,frame=lines, autogobble}

\usepackage{upgreek}
\usepackage{microtype}
\usepackage{url}

\newcommand{\model}{\texttt{bcNMF}\xspace}

\title{Disentangling Shared and Target-Enriched Topics via Background-Contrastive Non-negative Matrix Factorization}
\author{
Yixuan Li$^{1,2}$,
Archer Y. Yang$^{1,2,3,*}$,
Yue Li$^{2,3,*}$
}

\affil{
$^{1}$Department of Mathematics and Statistics, McGill University, Montreal, QC, Canada\\
$^{2}$Mila -- Quebec AI Institute, Montreal, QC, Canada\\
$^{3}$School of Computer Science, McGill University, Montreal, QC, Canada\\
$^{*}$Correspondence: archer.yang.yi@gmail.com; yueli@cs.mcgill.ca
}

\date{}

\begin{document}
\maketitle

\begin{abstract}
Biological signals of interest in high-dimensional data are often masked by dominant variation shared across conditions. This variation, arising from baseline biological structure or technical effects, can prevent standard dimensionality reduction methods from resolving condition-specific structure. The challenge is that these confounding topics are often unknown and mixed with biological signals.  Existing background correction methods are either unscalable to high dimensions or not interpretable. We introduce background contrastive Non-negative Matrix Factorization (\model), which extracts target-enriched latent topics by jointly factorizing a target dataset and a matched background using shared non-negative bases under a contrastive objective that suppresses background-expressed structure. This approach yields non-negative components that are directly interpretable at the feature level, and explicitly isolates target-specific variation. \model is learned by an efficient multiplicative update algorithm via matrix multiplication such that it is highly efficient on GPU hardware and scalable to big data via minibatch training akin to deep learning approach. Across simulations and diverse biological datasets, \model reveals signals obscured by conventional methods, including disease-associated programs in postmortem depressive brain single-cell RNA-seq, genotype-linked protein expression patterns in mice, treatment-specific transcriptional changes in leukemia, and TP53-dependent drug responses in cancer cell lines.
\end{abstract}

\section{Introduction}
High-dimensional data has become ubiquitous across biological and biomedical research, driven by advances in high-throughput measurement technologies, such as single-cell RNA sequencing, large-scale gene expression profiling, chromatin accessibility assays, multimodal single-cell platforms (e.g., CITE-seq) and electronic health records. These datasets routinely comprise thousands of features per sample and are used to study complex biological processes including cell identity, disease progression, and response to perturbation. A central challenge in analyzing such data is to extract latent structure that reflects biologically meaningful variation and is disentangled from technical bias and other sources of unwanted variation. In practice, biologically relevant signals are often weak relative to dominant background effects, making exploratory analysis and interpretation difficult \cite{stephens2017false, hicks2018missing}.

Dimensionality reduction is a critical step in the analysis of high-dimensional biological data, serving as the foundation for visualization, clustering, and downstream statistical modeling. Effective low-dimensional representations must balance two competing requirements: they should accurately capture structure relevant to the scientific question of interest, and they should remain interpretable at the level of individual genes, proteins, or clinical variables. Linear dimensionality reduction methods are particularly attractive in biological applications because they preserve additive structure and allow direct interpretation of feature loadings. As a result, principal component analysis (PCA) \cite{jolliffe2011principal} and non-negative matrix factorization (NMF) \cite{lee1999learning} have become standard tools across genomics and systems biology \cite{brunet2004metagenes, lehoucq1998arpack}. PCA identifies orthogonal directions that maximize explained variance in the data, providing compact summaries that are often effective for visualization and noise reduction. NMF, by contrast, constrains both basis vectors and coefficients to be non-negative, yielding additive, parts-based representations that are particularly well suited for gene expression, imaging, and count-based data \cite{lee1999learning, brunet2004metagenes}. These properties have made PCA and NMF foundational methods for exploratory analysis in biology. However, both approaches are fundamentally variance-driven: they prioritize directions or topics that explain the largest overall variation, regardless of whether that variation is biologically meaningful.

In many biological settings, the dominant sources of variance are not aligned with the signal of interest. Instead, leading topics often reflect technical artifacts, batch effects, cell-type composition, or baseline biological variability shared across experimental conditions. For example, in single-cell RNA sequencing (scRNA-seq) studies, PCA frequently separates cells by cell type while failing to resolve disease-associated transcriptional programs within a given cell type; similarly, in differential expression or perturbation studies, background heterogeneity can overwhelm subtle but systematic condition-specific effects \cite{abid2018exploring, anders2010differential}. As a consequence, standard dimensionality reduction methods can fail to recover rare or condition-specific patterns, even when those patterns are reproducible and biologically meaningful.

These limitations have motivated the development of contrastive learning approaches, which aim to isolate structure that is enriched in a target dataset relative to a reference or background dataset. Contrastive ideas appear broadly across machine learning and statistics, including classical hypothesis testing and likelihood-based comparisons, as well as noise-contrastive estimation, which frames representation learning as a discrimination task between observed data and a reference distribution \cite{gutmann2010noise}. More recently, modern contrastive objectives such as InfoNCE have become central to representation learning, where embeddings are learned by distinguishing foreground samples from negative or background examples \cite{oord2018representation}. While these methods have proven effective for learning transferable representations in vision, language, and multimodal settings \cite{chen2020simple, he2020momentum}, they are typically implemented using nonlinear encoders and implicit objectives. As a result, the learned representations are difficult to interpret at the feature level, limiting their suitability for exploratory biological analysis where transparency and direct attribution to genes or molecular features are essential.

In parallel, contrastive principles have been explicitly adapted to dimensionality reduction by comparing a target dataset against a background reference to suppress shared variation. Contrastive principal component analysis (cPCA) identifies linear directions that maximize variance in a target dataset while penalizing variance explained in a background dataset through a contrastive strength parameter $\alpha$ \cite{abid2018exploring}. While computationally efficient and easy to implement, cPCA is limited to linear structure, requires careful tuning of $\alpha$, and produces unconstrained topics that can be difficult to interpret for non-negative biological data. Probabilistic contrastive PCA (PCPCA) extends this framework within a linear Gaussian latent variable model, enabling likelihood-based inference and uncertainty quantification, but retains the need for hyperparameter tuning and Gaussian assumptions \cite{li2020probabilistic}. Related contrastive latent variable models explicitly decompose variation into shared and target-specific topics using probabilistic formulations, but remain sensitive to latent dimension choices and are typically restricted to Gaussian likelihoods \cite{severson2019unsupervised}.

To capture more complex structure, nonlinear contrastive generative models have been proposed. Contrastive variational autoencoders (cVAE) introduce separate latent spaces for salient and irrelevant variation using deep neural networks \cite{abid2019contrastive}. ContrastiveVI further specializes this approach for single-cell RNA sequencing by modeling shared and condition-specific latent variables under negative binomial or zero-inflated negative binomial likelihoods and incorporating batch covariates \cite{weinberger2023isolating}. Although effective in practice, these models are computationally intensive, require careful hyperparameter tuning, and yield representations that are difficult to interpret at the feature level. Recently, contrastive independent component analysis (cICA) extended independent component analysis to the contrastive setting by decomposing higher-order cumulant tensors of foreground and background data, providing increased expressivity and strong identifiability guarantees without assuming a Gaussian structure \cite{wang2025contrastive}. However, cICA relies on higher-order moment estimation and tensor decomposition, which can be computationally demanding and less naturally suited to count-based biological data. Collectively, these approaches demonstrate the value of explicitly modeling background structure, while highlighting persistent trade-offs among interpretability, modeling flexibility, and computational complexity.

In this study, we introduce background contrastive non-negative matrix factorization (\model), a contrastive dimensionality reduction method designed for interpretable comparative analysis of high-dimensional data. In \model, a target dataset and a matched background dataset are jointly factorized using a shared set of non-negative basis vectors, while maintaining separate coefficient matrices for each dataset. The method explicitly minimizes a contrastive objective that favors topics active in the target dataset while suppressing those expressed in the background. This formulation preserves the additive, parts-based structure of NMF, supports flexible divergence functions suitable for continuous and count-based data, and yields low-dimensional representations that are directly interpretable at the feature level.
We validate \model through a series of synthetic and real-world experiments. Using simulated image data with structured background confounding, we show that \model reliably recovers target-specific patterns where standard dimensionality reduction methods fail. We evaluate \model on several biological datasets, including protein expression data from mice, single-cell RNA-seq from leukemia patients, and large-scale perturbation data from cancer cell lines, where \model recovers condition-specific programs such as the canonical p53 transcriptional response while resolving dataset-specific heterogeneity. Finally, we apply \model to single-cell transcriptomic data from major depressive disorder (MDD) brain samples, demonstrating its ability to isolate disease-associated programs obscured by dominant background variation. Together, these results establish \model as a transparent and effective framework for contrastive dimensionality reduction in high-dimensional biological data.

\section{Results}

\begin{figure}[t!]
\centering\includegraphics[width=1.0\textwidth]{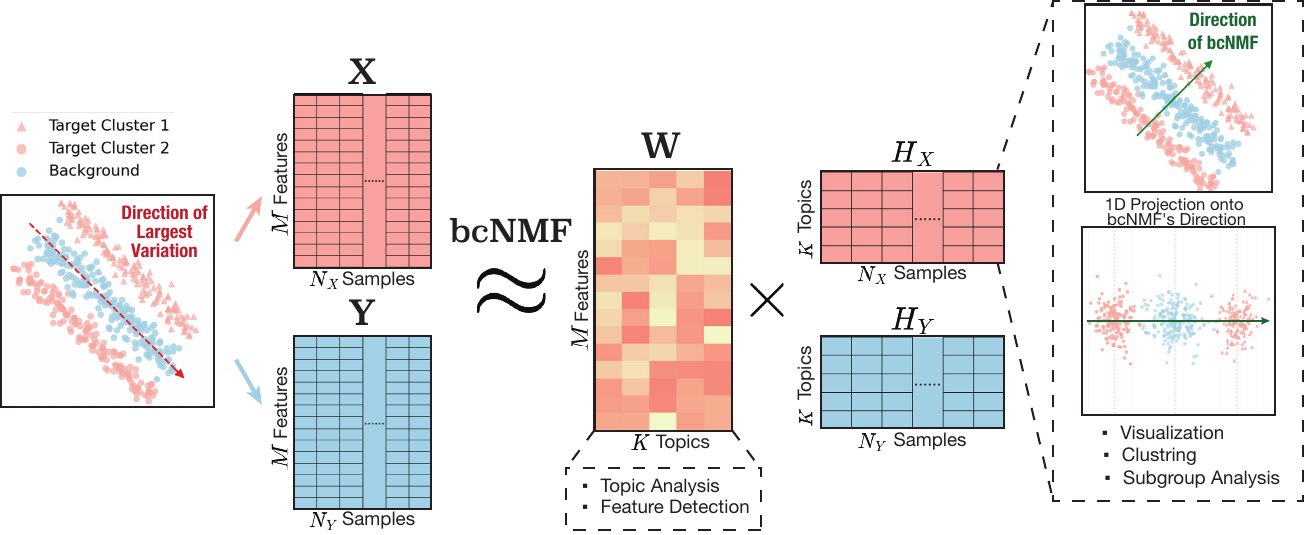}
\caption{Schematic overview of the \model framework. The figure illustrates the joint factorization of target and background datasets using \model, with both datasets decomposed into shared non-negative topics. Each sample is represented by its topic activations.}\label{fig:bcNMF_concept}
\end{figure}

\subsection{\model method overview}
Given a non-negative target data matrix $X \in \mathbb{R}_{\geq 0}^{M \times N_X}$, and a non-negative background data matrix $Y \in \mathbb{R}_{\geq 0}^{M \times N_Y}$, \model jointly factorizes both datasets using a shared non-negative basis matrix $W \in \mathbb{R}_{\geq 0}^{M \times K}$, together with dataset-specific coefficient matrices $H_X \in \mathbb{R}_{\geq 0}^{K \times N_X}$ and $H_Y \in \mathbb{R}_{\geq 0}^{K \times N_Y}$ (Fig.~\ref{fig:bcNMF_concept}). Here, $M$ corresponds to features (e.g., genes, proteins, or pixels), $N_X$ and $N_Y$ correspond to samples, and $K$ denotes the number of latent topics used to represent the data. This formulation enforces a common set of latent topics across datasets while allowing their activations to differ between the target and the background. The factorization is obtained by minimizing a contrastive objective function:
$$
\min _{W, H_X, H_Y \geq 0} \mathcal{L}\left(X, W H_X\right)-\alpha \mathcal{L}\left(Y, W H_Y\right),
$$
where $\mathcal{L}(\cdot, \cdot)$ denotes a loss function corresponding to an appropriate negative data likelihood, and $\alpha$ controls the strength of background suppression. Depending on the data modality, $\mathcal{L}$ may represent a squared reconstruction loss (Gaussian), a negative Poisson log-likelihood, or a negative (zero-inflated) negative binomial log-likelihood. By jointly encouraging accurate reconstruction of the target data and suppressing reconstruction of the background, \model emphasizes latent topics that are specific to $X$ relative to $Y$. Unlike the standard NMF, which optimizes reconstruction of a single dataset, \model explicitly incorporates a background reference into the objective, enabling direct suppression of the shared background structure during factorization. For scalability to large datasets, we also optimize the model using mini-batch training~\ref{sec:app_minibatch}.

\begin{figure}[t!]
\centering\includegraphics[width=1.0\textwidth]{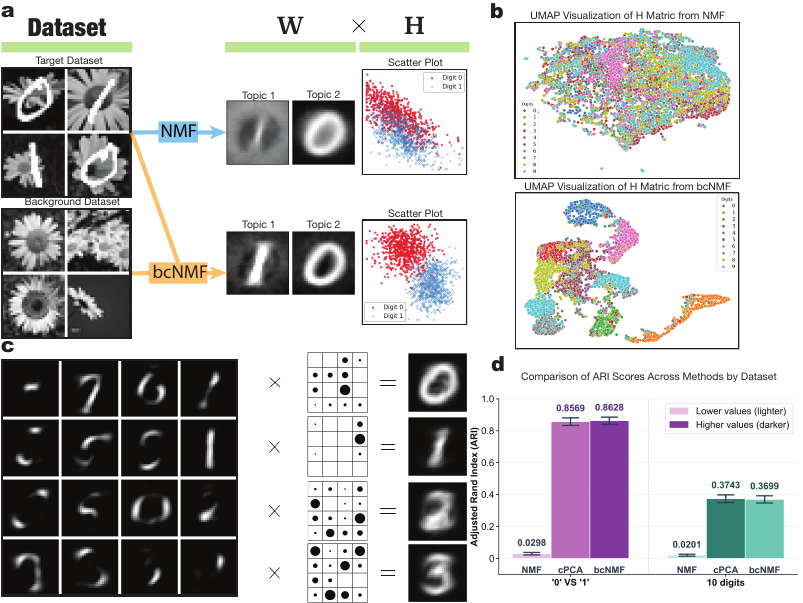}
\caption{Simulation results on blended MNIST-ImageNet data.
\textbf{a} Schematic illustrating the construction of the target and background datasets. The target dataset comprises MNIST digits superimposed on ImageNet background patches, while the background dataset contains only natural images processed identically.
\textbf{b} UMAP visualization of the coefficient matrix $H_X$ learned from standard NMF (top) and \model (bottom) on the ten-digit dataset. 
\textbf{c} Interpretability of \model topics. Each column of the basis matrix $W$ is visualized as a $28 \times 28$ image (left), with topic usage for each digit represented as bubble plots (center). The mean reconstructed image for each digit, obtained by combining $W$ with the corresponding average topic vector, is shown on the right.
\textbf{d} Quantitative comparison of clustering performance. ARI scores for NMF, cPCA, and \model on the binary and ten-digit tasks, with error bars indicating bootstrap standard errors estimated from repeated subsampling of the test data.
}\label{fig:MNIST}
\end{figure}

\subsection{Evaluation on simulated image data}
To assess the ability of \model to recover target-specific structure under strong background confounding, we constructed a simulation using handwritten digit images from MNIST \cite{lecun1998mnist} and natural images from ImageNet \cite{deng2009imagenet}. We focused initially on digits 0 and 1, sampling approximately one tenth of the MNIST training set in a stratified manner, yielding 690 images of digit 0 and 787 images of digit 1 (Fig. ~\ref{fig:MNIST}a). Each image was resized to $28 \times 28$ pixels and reshaped into a vector of length 784. As background, we sampled 650 grayscale images from a single ImageNet category (flowers), resized and center-cropped to the same resolution. Using a single ImageNet category ensured that background variation dominated the overall variance while maintaining coherent structure across background samples.

Target images were generated by superimposing a randomly selected MNIST digit onto a randomly chosen ImageNet background patch. The composite image was formed using an element-wise maximum of the digit and background, with background pixel intensities scaled by a factor of 0.8 to preserve visible digit structure while introducing substantial background variation. This construction produces images in which foreground and background features are both prominent. Pixel values were normalized to the range $[0,1]$. The resulting target dataset consists of digit images embedded in structured natural backgrounds (Fig.~\ref{fig:MNIST}a, top), while the background dataset comprises ImageNet images processed identically but without superimposed digits (Fig.~\ref{fig:MNIST}a, bottom).

We applied standard NMF and our \model to the target and background datasets using $K=2$ topics. For both methods, we plotted the target coefficients $H_X$ directly in two dimensions, with each point representing a sample and coordinates given by its two topic coefficients. In the NMF embedding, samples corresponding to digits 0 and 1 overlap extensively, forming a single diffuse cluster (Fig.~\ref{fig:MNIST}a, right). In contrast, \model yields two well-separated groups aligned with the true digit labels, indicating that incorporation of background information enables recovery of digit identity despite strong background variation.

We next extended the simulation to all ten MNIST digits. In this setting, the target dataset contains digits 0-9, each composited with an ImageNet background as above. We used $K=16$ topics for both NMF and \model to enable flexible representations, where  individual topics capture substructures rather than entire digits. To visualize the resulting embeddings, we projected the coefficient matrix $H_X$ into two dimensions using UMAP with default parameters and a fixed random seed (Fig.~\ref{fig:MNIST}b). In the NMF embedding, samples corresponding to different digits largely overlap. By contrast, \model produces visually distinct clusters corresponding to individual digits, consistent with improved separation in the latent space.

Beyond clustering performance, \model preserves the interpretability of NMF. Each column of the basis matrix $W$ corresponds to a spatial pattern that can be reshaped into a $28 \times 28$ image. To summarize how digits are composed of these patterns, we computed the mean coefficient vector for each digit class and normalized it to sum to one, yielding topic proportions. These compositions are visualized in Fig.~\ref{fig:MNIST}c (center), and multiplication of the mean topic vectors with $W$ reconstructs representative images for each digit (Fig.~\ref{fig:MNIST}c, right), confirming that distinct digits arise from different combinations of shared latent patterns.

To quantitatively evaluate clustering performance, we computed the adjusted Rand index (ARI) \cite{hubert1985comparing} between predicted clusters and true digit labels for both the two-digit and ten-digit settings (Fig.~\ref{fig:MNIST}d). Clusters were obtained using k-means applied to the coefficient matrix $H_X$, with $K=2$ or $K=10$ as appropriate, and all methods were initialized randomly. For \model, the contrastive parameter $\alpha$ was selected to minimize reconstruction error. In the binary task, cPCA and \model achieved high ARI values (0.8569 and 0.8628, respectively), whereas NMF performed poorly (ARI = 0.0298). In the ten-digit setting, cPCA and \model again substantially outperformed NMF (ARI = 0.3743 and 0.3699 versus 0.0201).  
The error bars are small relative to the gap between NMF and the contrastive methods in both tasks, indicating that the near-random performance of NMF and the strong separation achieved by cPCA and \model are stable under data resampling.

\begin{figure}[t!]
\centering\includegraphics[width=0.6\textwidth]{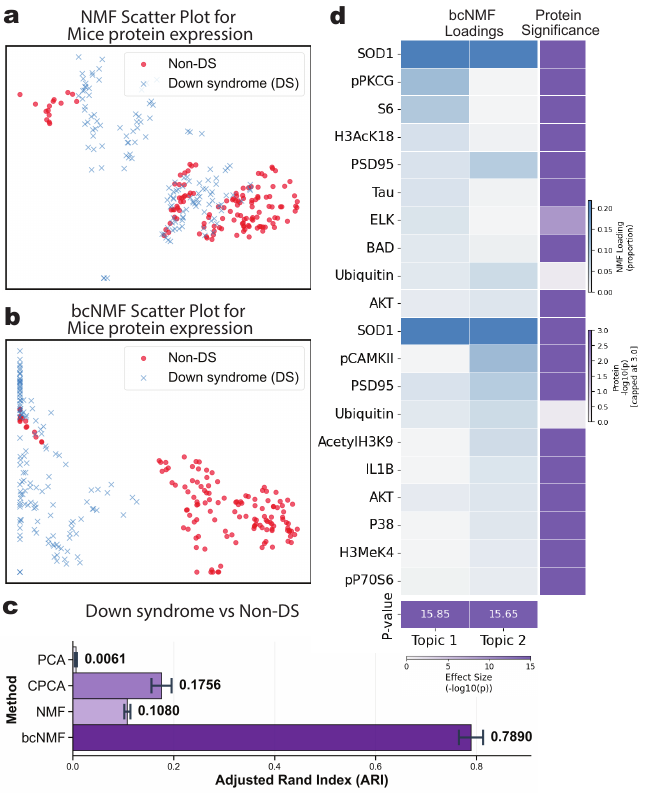}
\caption{Down syndrome-associated protein expression in mice.
\textbf{a} UMAP embedding of NMF topic usage for mouse protein expression data, colored by Down syndrome (DS) status.
\textbf{b}  UMAP embedding of \model topic usage using control mice as background, colored by DS status.
\textbf{c}  Bar plot of ARI for DS classification using PCA, cPCA, NMF and \model. Error bars indicate bootstrap standard errors estimated from 20 subsamples with replacement (300 target and 200 background samples per subsample).
\textbf{d}  Top ten proteins for each of the two most significant \model topics. Heatmaps show protein loadings (left) and signed $-\log_{10} p$ values from Welch’s two-sample t-tests comparing DS and non-DS groups (right).}\label{fig:Mice}
\end{figure}

\subsection{Unraveling protein expression topics specific to Down syndrome}

We next evaluated \model on a protein expression dataset derived from mice subjected to shock therapy \cite{sarkar2008proteomic, geddes1990cases}. The target dataset consists of protein abundance measurements from shocked mice, including both mice carrying the Down syndrome (DS) genotype and non-DS mice. The DS status was treated as hidden ground truth, withheld during the analysis, and used only for post hoc evaluation. We seek to assess whether latent variation within the shocked mice population can be identified in an unsupervised manner and whether this variation distinguishes DS from non-DS mice.

We initially applied NMF with $K = 2$ topics to the shocked mice dataset and visualized the corresponding coefficient matrix $H$, with each data point representing an individual mouse (Fig.~\ref{fig:Mice}a). DS and non-DS mice exhibited substantial overlap, with no clear separation between subgroups. This suggests that the major sources of variation among the shocked mice are likely driven by natural biological differences, such as sex or age, rather than DS genotype.

We then used a background dataset consisting of protein expression measurements from a separate set of control mice that had not been exposed to shock therapy. These control mice are expected to share similar sources of natural biological variability (e.g.,  age, sex, baseline physiology). We applied \model with $K=2$ topics using both the target and background datasets and visualized the corresponding coefficient matrix of the target data ($H_X$), with each data point representing an individual mouse (Fig.~\ref{fig:Mice}b). In contrast to standard NMF, the \model embedding showed reduced overlap between DS and non-DS mice, indicating improved separation of latent structure associated with DS status when background information is incorporated.

To quantitatively assess the subgroup separation, we clustered the topic coefficients using k-means with $K=2$ and computed the ARI with respect to the withheld DS labels (Fig.~\ref{fig:Mice}c). PCA showed negligible alignment with DS status (ARI = 0.0061), while cPCA (ARI = 0.176) and NMF (ARI = 0.108) achieved modest improvements. \model attained a substantially higher ARI of 0.789, indicating stronger concordance between the unsupervised clusters and DS genotype under this evaluation protocol. 

To investigate the molecular content of the latent dimensions identified by \model, we examined the distribution of the two topics and their association with DS (Fig.~\ref{fig:Mice}d). For each topic, we selected the top ten proteins with the largest topic scores. At the topic level, association with DS status was evaluated using logistic regression of the binary DS label on topic usage $H_X$ across samples, yielding $-\log_{10}p$ values of 15.85 and 15.65 for the two topics. 

We also assessed associations between individual protein abundances and DS status using univariate logistic regression at the sample level. Many of the high-loading proteins are also significantly associated with DS status (Fig.~\ref{fig:Mice}d). Notably, SOD1 appeared among the top contributors to both topics, consistent with its established role in DS-related oxidative stress. Additional proteins with strong loadings include synaptic and signaling regulators such as PSD95 and pCAMKII, as well as regulators of growth, stress, and apoptotic pathways including S6, AKT, p38, and IL-1$\beta$. Together, these results indicate that \model captures latent protein expression patterns that are coherently associated with DS genotype and align with known biological features of Down syndrome.

\begin{figure}[b!]
\centering\includegraphics[width=0.85\textwidth]{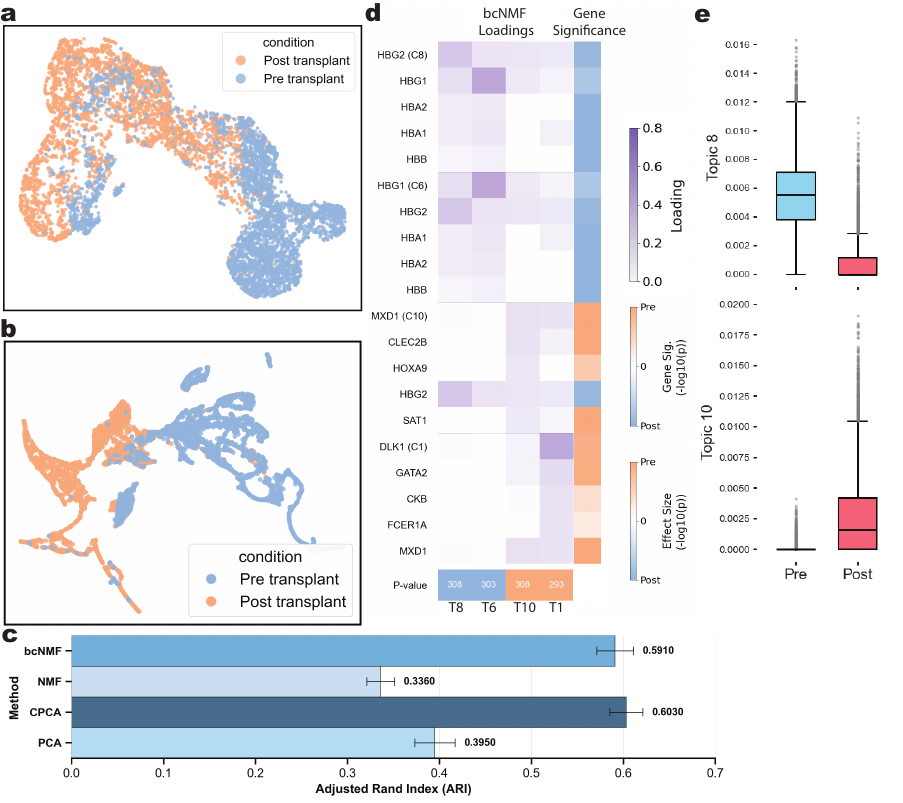}
\caption{\model resolves condition-specific programs in single-cell RNA-seq from a leukemia transplant patient.
\textbf{a} UMAP embedding of NMF topic usage ($K$ = 10), colored by condition (pre-transplant, blue; post-transplant, orange).
\textbf{b} UMAP embedding of \model topic usage using healthy bone marrow as background, colored by condition.
\textbf{c} ARI for transplant status classification using PCA, cPCA, NMF and \model. Error bars denote bootstrap standard errors from 20 stratified subsamples ($n$ = 6,000 cells per subsample).
\textbf{d} Top five genes for each of the four most significant \model topics. Heatmaps show gene loadings (left) and signed $-\log_{10} p$ values from Welch’s two-sample t-tests comparing pre- and post-transplant cells (right).
\textbf{e} Box plots of topic usage for two \model topics (Topics 8 and 10) in pre- and post-transplant cells; each point represents a single cell.}\label{fig:Leukemia}
\end{figure}

\subsection{Identifying stem cell transplantation effects in leukemia}

We next applied \model to an scRNA-seq dataset generated from bone marrow mononuclear cells (BMMCs) collected from a leukemia patient before and after stem cell transplantation, enabling within-patient comparison of treatment-associated transcriptional changes \cite{zheng2017massively}. As a background reference, we incorporated scRNA-seq data from healthy donor BMMCs derived from the same 10x Genomics platform, thereby capturing baseline hematopoietic heterogeneity independent of treatment. The healthy background dataset comprised approximately 12,000 cells from a single donor. No explicit batch correction was performed, as both patient samples and background samples were generated using comparable experimental protocols. We chose the top 3,000 highly variable protein-coding genes for all downstream factorization analyses.

We applied standard NMF and \model with $K=10$ topics to the leukemia data. The resulting lower dimensional representations ($H$ for NMF and $H_X$ for \model) were projected using UMAP with default parameters and a fixed random seed (Fig.~\ref{fig:Leukemia}a,b). In the NMF embedding (Fig.~\ref{fig:Leukemia}a), pre- and post-transplant cells exhibited substantial overlap and were distributed along partially intermingled manifolds. In contrast, the \model embedding (Fig.~\ref{fig:Leukemia}b) showed reduced overlap and visually clearer separation between pre- and post-treatments. This difference is consistent with the contrastive objective suppressing variation shared with healthy background samples, thereby revealing treatment-specific gene expression programs.
We then quantified the separation between conditions by the ARI of k-means clustering ($K=2$) with respect to the transplant status (Fig.~\ref{fig:Leukemia}c). NMF showed the lowest agreement with the transplant status (ARI = 0.336) trailing PCA (ARI = 0.395). Both cPCA (ARI = 0.603) and \model (ARI = 0.591) achieved higher alignment than PCA and NMF. 

We identified topics associated with the treatment using Welch’s two-sample t-tests comparing topic usage between pre- and post-transplant cells. Four topics exhibit significant condition specificity: two enriched in pre-transplant cells (Topics 8 and 6) and two enriched in post-transplant cells (Topics 10 and 1) (Fig.~\ref{fig:Leukemia}d). For each topic, the top genes exhibited concordant directionality and condition-specific enrichment, indicating consistency between topic-level structure and gene-level association.
Notably, Topic 8 showed higher usage in pre-transplant cells, whereas Topic 10 was elevated in post-transplant cells (Fig.~\ref{fig:Leukemia}e). The observed shifts in usage distributions were consistent across large numbers of cells and aligned with the directionality indicated by the t-tests. 
The pre-transplant topics (Topics 8 and 6) were dominated by hemoglobin genes including \textit{HBG2}, \textit{HBG1}, \textit{HBA2}, \textit{HBA1}, and \textit{HBB}, consistent with an erythroid-associated transcriptional program. While hemoglobin expression can reflect erythroid lineage activity, we note that single-cell RNA-seq datasets may also capture erythroid-related signals arising from ambient RNA or lineage composition shifts. In contrast, post-transplant topics (Topics 10 and 1) were characterized by genes such as \textit{MXD1}, \textit{CLEC2B}, \textit{SAT1}, \textit{DLK1}, and \textit{GATA2}, suggesting altered differentiation and immune-related programs following transplantation. These interpretations are based on gene-level enrichment patterns and should be viewed as hypothesis-generating rather than definitive functional assignments.

Collectively, these analyses indicate that \model identifies two opposing, condition-aligned transcriptional programs—one preferentially enriched prior to transplantation and one enriched afterward—while retaining gene-level interpretability and improving resolution of treatment-associated variation in the latent representation.

\begin{figure}[t!]
\centering\includegraphics[width=\textwidth]{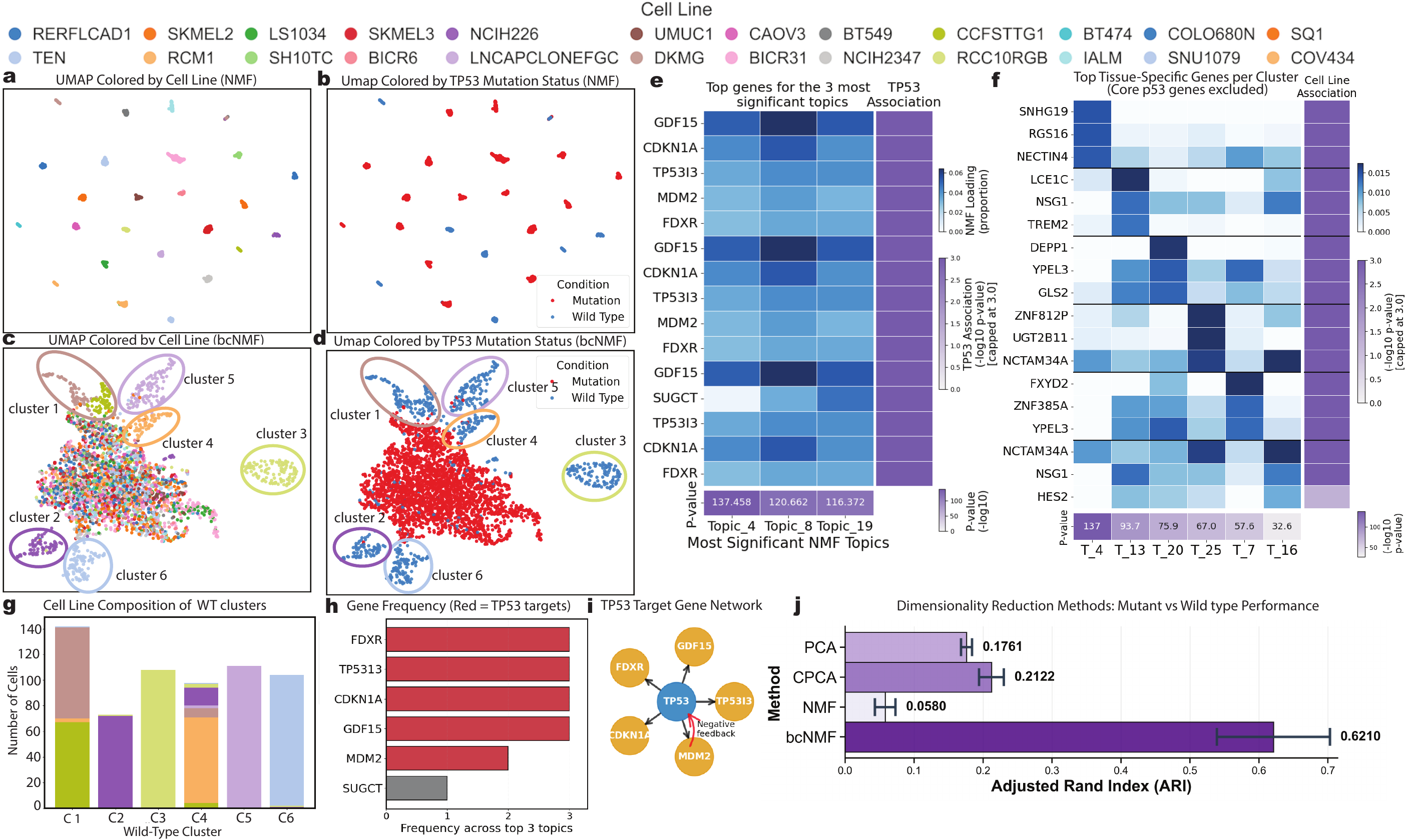}
\caption{\model isolates TP53-dependent transcriptional responses in the MIX-seq idasanutlin dataset.
\textbf{a} UMAP embedding of NMF topic usage (K = 20), colored by cell line identity.
\textbf{b}  UMAP embedding of NMF topic usage, colored by TP53 mutation status (wild-type, mutant).
\textbf{c}  UMAP embedding of \model topic usage (K = 20) using wild-type cell lines as background, colored by cell line identity.
\textbf{d}  UMAP embedding of \model topic usage, colored by TP53 mutation status.
\textbf{e}  Top five genes for each of the three most significant \model topics. Heatmaps show gene loadings (left) and signed $-\log_{10} p$ values from Welch’s two-sample t-tests comparing TP53 mutant and wild-type cells (right); bottom bars show topic-level significance.
\textbf{f}  Cell-line-specific genes from \model topics enriched in wild-type clusters, excluding canonical TP53 targets. Heatmaps show gene loadings and Welch’s t-test-based significance scores.
\textbf{g}  Histogram of cell line composition within wild-type TP53 clusters identified in the \model embedding.
\textbf{h}  Frequency of genes identified across the three most significant \model topics 4, 8 and 19, with canonical TP53 targets highlighted in red.
\textbf{i}  Schematic of the TP53 regulatory network linking identified target genes.
\textbf{j} ARI for TP53 mutation status alignment using PCA, cPCA, NMF and \model. Bars show ARI computed on the full dataset; error bars indicate bootstrap standard errors estimated from 20 resampled target/background pairs
}\label{fig:Mcfarland_result}
\end{figure}

\subsection{Unraveling drug perturbation effects on cancer transcriptome}

We further showcase \model by analyzing an scRNA-seq perturbation dataset \cite{mcfarland2020multiplexed}. The dataset comprised 3,771 cells from 24 cancer cell lines treated with idasanutlin, an MDM2 antagonist that activates the tumour-suppressor protein p53, in TP53 wild-type cell lines (that is, cells with an intact TP53 gene). In parallel, 3,439 cells were treated for 24 hours with DMSO (dimethyl sulfoxide), which served as the vehicle control. Idasanutlin-treated cells were designated as the target dataset and DMSO-treated cells as the background dataset. Among the 24 cell lines, seven carried wild-type TP53 and were expected to exhibit transcriptional response to p53 activation, whereas the remaining 17 cell lines carried mutant TP53 and considered nonresponsive. All samples were generated using the same MIX-seq experimental platform and protocol. No explicit batch correction was performed, as both treatment conditions were processed under comparable experimental settings. We focused our analysis on the top 3,000 HVGs computed across the combined target and background matrices. 

For all factorization approaches, we learned latent representations using 
$K=20$ topics. In the NMF embedding (Fig.~\ref{fig:Mcfarland_result}a), we observed that cells segregated into 24 well-separated clusters when colored by cell line identity, demonstrating that baseline transcriptional differences among cell lines dominated the factorization. When we colored the same embedding by TP53 mutation status (Fig.~\ref{fig:Mcfarland_result}b), wild-type and mutant cells remained organized according to cell line structure, with no evident genotype-driven grouping. These results show that standard NMF primarily captures baseline cell line heterogeneity, and mutation status does not emerge as a major organizing axis of the latent space.

In contrast, \model produced a substantially different organization of the data. When we colored the \model embedding by cell lines (Fig.~\ref{fig:Mcfarland_result}c), several lines formed distinct peripheral clusters (e.g., clusters 2, 3, 5, and 6), while many others were intermingled within a large central manifold. This pattern indicates that \model attenuates dominant baseline separation across cell lines while retaining distinct structure for strongly divergent lines. When we colored the same embedding by TP53 mutation status (Fig.~\ref{fig:Mcfarland_result}d), we observed clearer genotype-dependent organization: cells from wild-type TP53 lines occupied multiple coherent clusters, whereas mutant lines largely populated a broad, overlapping region. Examination of cluster composition (Fig.~\ref{fig:Mcfarland_result}g) further showed that, except for cluster 1, each wild-type cluster was dominated by a single responsive cell line with minimal cross-line mixing. Together, these results demonstrate that \model suppresses dominant baseline heterogeneity and reorganizes the latent space to better align treatment-responsive genotypes while preserving within-line identity.

We quantified associations between latent topics and TP53 mutation status by performing univariate logistic regression of the binary TP53 genotype on normalized topic usage. Significance was summarized as signed $-\log_{10} p$ values. The three most strongly associated topics were selected for detailed inspection (Fig.~\ref{fig:Mcfarland_result}e). For each topic, we ranked the top five genes by loading and visualized them alongside their signed $-\log_{10} p$ values, computed from independent univariate logistic regressions applied directly to the original gene expression matrix. Across the three topics, a highly overlapping set of genes emerged, including MDM2, CDKN1A, TP53I3, FDXR, and GDF15, all canonical transcriptional targets of p53. The recurrence of these genes across topics reflects the specificity and strength of the TP53-dependent transcriptional response to idasanutlin. Gene recurrence frequency across the top three topics is summarized in Fig.~\ref{fig:Mcfarland_result}h where we highlight the majority of genes that appeared in all topics and are canonical TP53 targets.  We illustrated the regulatory relationships among these genes in Fig.~\ref{fig:Mcfarland_result}i, consistent with established p53 pathway biology \cite{vousden2009blinded, riley2008transcriptional}. The concordance between topic loadings and independent gene-level regressions confirms that the model-derived structure reflects reproducible signal in the raw expression space.

Heterogeneity among wild-type clusters remained evident even after alignment by TP53 status. To reveal cluster-specific programs beyond the dominant p53 response, canonical TP53 target genes were excluded, as these genes strongly drive multiple contrastive topics and obscure secondary structure. The resulting gene set (Fig.~\ref{fig:Mcfarland_result}f) uncovers cluster-specific lineage signatures that differentiate wild-type cell populations.
The heatmap shows that each cluster is characterized by a small set of selectively enriched genes with minimal overlap across clusters. For example, NSG1 and RGS16 are strongly enriched in one cluster, whereas NECTIN4 and DEPP1 define another, and GLS2 and LNCTAM34A mark additional clusters. These genes exhibit pronounced cell line-specific association signals, consistent across both topic loadings and expression-level statistics. This pattern demonstrates that \model isolates a shared TP53 activation program while simultaneously preserving layered, cell line-specific transcriptional modules. Rather than collapsing all responsive lines into a uniform genotype-driven state, the contrastive representation retains intrinsic lineage identity on top of the common p53 response, revealing structured heterogeneity within the wild-type group.

We then quantified the separation between wild-type and mutant TP53 lines using the ARI between $k$-means clusters ($K=2$) derived from each low-dimensional representation and TP53 mutation status. 
As shown in Fig.~\ref{fig:Mcfarland_result}j, NMF exhibited minimal alignment with TP53 mutation status (ARI = 0.058), indicating that baseline cell line heterogeneity dominated the embedding. PCA (ARI = 0.176) and cPCA (ARI = 0.212) showed moderate improvements but remained limited in resolving genotype structure. In contrast, \model achieved substantially higher agreement (ARI = 0.621), demonstrating markedly improved and stable genotype separation relative to the other methods. Notably, the bootstrap error bars do not overlap with those of the alternative methods, further supporting the robustness and statistical separation of \model’s performance.

Collectively, these results demonstrate that \model isolates TP53-dependent transcriptional responses in a setting where strong baseline cell line heterogeneity would otherwise dominate unsupervised structure. \model simultaneously preserves intrinsic cell line identity, recapitulates the canonical p53 transcriptional program, and resolves additional line-specific transcriptional modules layered on top of the shared drug response.

\begin{figure}[t!]
\centering\includegraphics[width=1.0\textwidth]{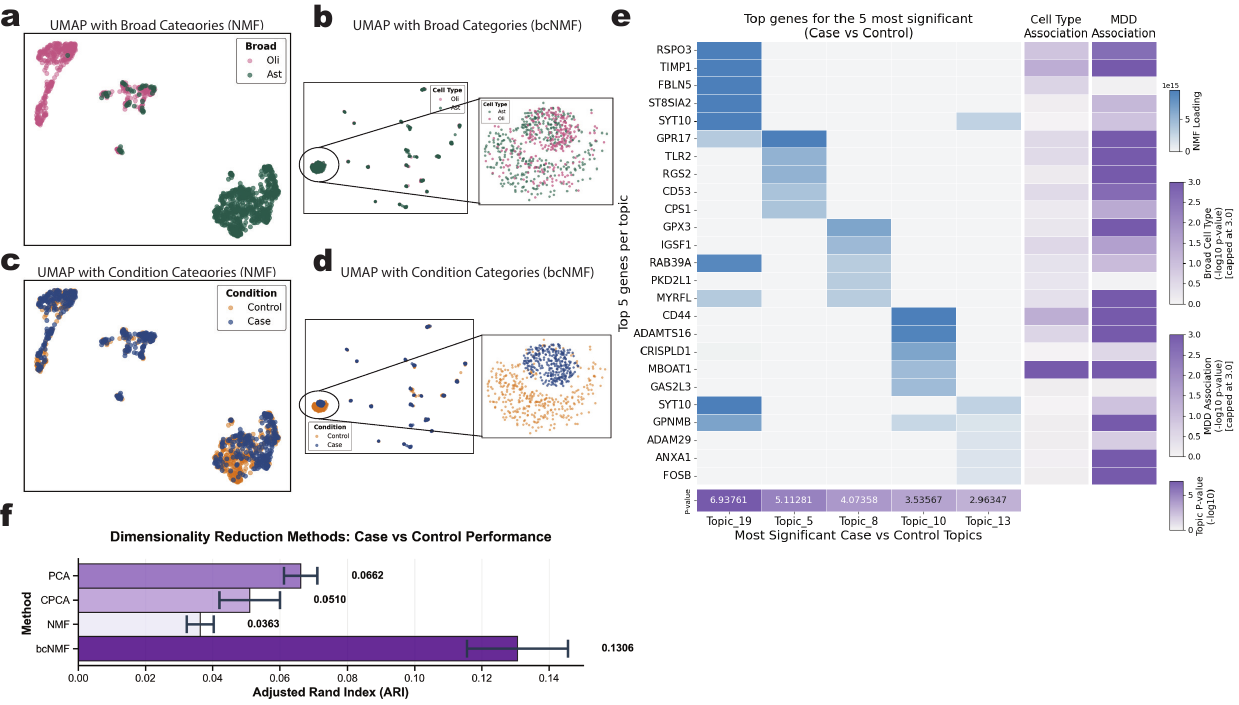}
\caption{Revealing hidden disease-associated programs in single-cell RNA-seq from MDD.
\textbf{a} UMAP embedding of NMF topics colored by cell types (astrocyte, oligodendrocyte).
\textbf{b} UMAP embedding of \model topics colored by cell types.
\textbf{c} UMAP embedding of NMF topics colored by condition (case, control).
\textbf{d} UMAP embedding of \model topics colored by condition.
\textbf{e} Top five genes for each of the five most significant \model case-control topics (y-axis), with heatmap showing gene loadings and topic-wise regression p-values (bottom row) and MDD-associated pathway enrichment scores (right panel). 
\textbf{f} Bar plot of ARI for case-control clustering using PCA, cPCA, NMF, and \model. Higher ARI indicates better recovery of disease-relevant groupings. Error bars were estimated by bootstrap with 20 independent resamples drawing 6,000 cells from the target dataset and 3,000 cells from the background dataset with replacement.}\label{fig:MDD_result1}
\end{figure}

\subsection{Disentangling transcriptome markers for depression from cell type functions}

As an exploratory analysis, we applied \model to a single-nucleus RNA-seq dataset generated from postmortem human dorsolateral prefrontal cortex (dlPFC) \cite{maitra2023cell}. The full dataset comprises over 160,000 nuclei from individuals with major depressive disorder (MDD) and matched controls. To reduce demographic and technical heterogeneity, we restricted our analyses to female donors aged 40-60 years of Caucasian descent from the Douglas Bell-Canada Brain Bank. After filtering, 11 MDD cases and 4 controls remained. For balanced case-control comparison, four MDD cases were randomly selected (fixed random seed) and paired with the four available controls. Repeated random selections yielded qualitatively consistent results. Given prior evidence that cell-type identity constitutes the dominant axis of variation in cortical single-cell transcriptomic data, and that glial populations are repeatedly implicated in MDD pathology in the original study \cite{maitra2023cell}, we further restricted primary analyses to astrocytes (Ast) and oligodendrocytes (Oli). Focusing on these two glial populations serves two purposes: it mitigates confounding driven by broad neuronal-glial differences and concentrates the analysis on cell types with established disease relevance. The target dataset therefore consisted of astrocyte and oligodendrocyte cells from the four selected MDD cases and four controls, whereas the background dataset consisted of the same four control samples only. This design allows disease-associated variation to remain present within the target dataset while enabling suppression of variation shared with controls. Raw count matrices were analyzed without log transformation. Gene selection proceeded by restricting to protein-coding genes (26,639 reduced to 18,716 genes), followed by selection of the top 3,000 HVGs computed across the combined target and background matrices using Scanpy \cite{wolf2018scanpy}. Cell-type filtering was performed after HVG selection to avoid biasing variability estimates. The final matrices contained 8,886 target cells and 3,645 background cells, each measured across 3,000 HVGs. No additional batch correction was performed, as donors were selected within a constrained demographic subset and processed under a consistent experimental protocol in the original study.

In the standard NMF embeddings, astrocytes and oligodendrocytes formed clearly separated clusters (Fig.~\ref{fig:MDD_result1}a), confirming that cell-type identity constitutes the dominant axis of variation. Moreover, visualizing by diagnostic status revealed no clear separation between MDD cases and controls (Fig.~\ref{fig:MDD_result1}c), indicating that disease-associated signal is weak relative to cell-type structure under non-contrastive factorization. Applying \model with a Poisson likelihood substantially altered the geometry of the latent space. The contrastive parameter $\alpha$ was selected using model-intrinsic criteria based on reconstruction contrast and stability across a moderate range of values. In the \model embedding, astrocytes and oligodendrocytes appeared more intermixed (Fig.~\ref{fig:MDD_result1}b), consistent with attenuation of variation shared between target and background datasets, including dominant cell-type signals. More importantly, the \model embedding revealed a visibly sharper separation between MDD cases and controls (Fig.~\ref{fig:MDD_result1}d). Notably, a subset of cells forms a distinct case-enriched region that is magnified in the inset (Fig.~\ref{fig:MDD_result1}d), where cases cluster densely and are spatially separated from controls. This localized structure is not apparent in the NMF embedding and suggests that contrastive modeling exposes a disease-aligned subpopulation that is otherwise masked by the shared biological variation. We quantified the case-control separation in terms of ARI based on the $k$-means clusters ($K=2$) derived from each low-dimensional representation. 
Under this evaluation, \model achieved the highest ARI (0.1306), exceeding PCA (0.0662), cPCA (0.0510), and NMF (0.0363). Although absolute ARI values remain modest, consistent with subtle disease effects in postmortem cortical tissue, the improvement relative to non-contrastive approaches indicates enhanced recovery of disease-associated structure following suppression of shared cell-type variation. Together, these patterns indicate that \model suppresses dominant shared structure while enhancing condition-enriched transcriptional programs in the latent space.

We quantified the topic-disease associations based on the trained \model with $K=20$ topics. 
Five topics exhibited the strongest associations (Fig.~\ref{fig:MDD_result1}e). We extracted the top five genes from each of these topics, yielding a 25-gene signature panel (Fig.~\ref{fig:MDD_result1}e, heatmap). Enrichment analysis using Enrichr \cite{chen2013enrichr} with the 3,000 HVGs as background shows that these 25 genes were enriched for mitochondrial organization, oxidative stress, immune signaling, and cytokine-related pathways (Fig.~\ref{fig:MDD_app}).
Moreover, extracellular matrix (ECM) organization and neuroinflammatory processes are among the most enriched gene sets. Specifically, ECM organization and degradation pathways exhibited the highest gene ratios (0.35-0.45). GO immune and inflammatory terms showed moderate gene ratios (0.20-0.33), suggesting coordinated activation of neuroimmune programs. Network analysis of the top 12 enriched GO Biological Process terms (Fig.~\ref{fig:MDD_app}b) demonstrated a densely connected immune-inflammatory module with high gene overlap (Jaccard similarity \cite{niwattanakul2013using} $>$ 0.2), alongside peripheral metabolic terms with limited connectivity. The resulting topology (clustering coefficient = 0.42 versus random expectation $\approx$ 0.15) indicates that the identified genes participate in functionally integrated pathways. Together, these analyses suggest that contrastive topics capture coherent programs involving ECM remodeling and neuroinflammatory signaling, both previously implicated in MDD pathology.

\subsection{Runtime benchmark}

\begin{figure}[b!]
\centering\includegraphics[width=0.9\textwidth]{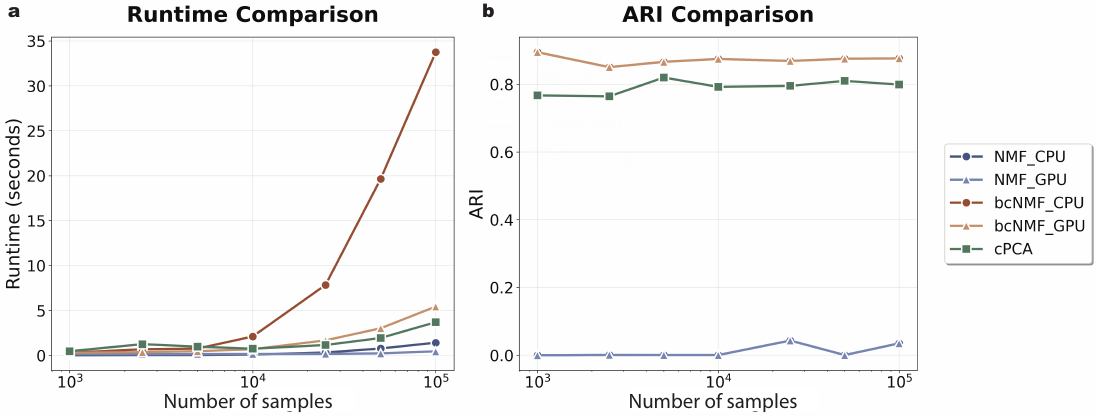}
\caption{Runtime and clustering performance comparison.
\textbf{a} Runtime of NMF and \model implemented on CPU (NumPy) and GPU (PyTorch) as a function of the number of MNIST-ImageNet image samples.
\textbf{b} Adjusted Rand Index (ARI) between ground-truth labels and clusters obtained from k-means applied to low-dimensional representations from PCA, cPCA, NMF, and \model.}\label{fig:runtime}
\end{figure}

To evaluate computational efficiency and scalability across methods, we benchmarked runtime performance as a function of sample sizes. We used the MNIST-ImageNet composite dataset described in Section 2.2, systematically varying the number of samples from 1,000 to 100,000 while holding feature dimensionality fixed at 784. For each sample size, we measured wall-clock time for each method to complete factorization or dimensionality reduction. For \model and NMF, we implemented both CPU (NumPy) and GPU (PyTorch with MPS backend) versions to assess hardware acceleration benefits. Both NMF and \model were run until convergence using a relative change tolerance of 0.01\% in reconstruction loss, with maximum iterations set to 500. For cPCA, we used the original Python implementation \cite{abid2018exploring}. All methods used $K = 2$ topics for consistency and the experiments were performed on a system with an Apple M1 Max processor and 32 GB of unified memory.

NMF exhibited the fastest performance on both CPU and GPU, scaling approximately linearly with sample size (Fig.~\ref{fig:runtime}a). At 1,000 samples, NMF completed in under 0.1 seconds on both devices, increasing to approximately 1 second (CPU) and 0.1 seconds (GPU) at 100,000 samples. \model required longer runtime, primarily because it required more iterations to reach the same convergence criterion due to the additional contrastive term in the objective, taking approximately 0.5 seconds (CPU) and 0.1 seconds (GPU) at 1,000 samples, and  33 seconds (CPU) and 5 seconds (GPU) at 100,000 samples. GPU acceleration provided consistent speedups of approximately 5--7 times for both NMF and \model across all scales. cPCA, implemented as a closed-form eigendecomposition of the contrastive covariance matrix, scaled more favorably than \model on CPU, completing in under 10 seconds across all tested sample sizes.

Clustering performance remained stable across sample sizes for all methods (Fig.~\ref{fig:runtime}b). \model and cPCA both maintained high ARI values above 0.8 across all scales, confirming that performance gains from contrastive learning were robust to dataset size. Notably, \model consistently outperformed cPCA by approximately 0.1 ARI across all tests, suggesting that the non-negative factorization framework more effectively captured the additive structure of digit composition. NMF consistently produced near-zero ARI values, reflecting its inability to resolve digit structure in the presence of dominant background variation. These results indicate that \model achieves competitive runtime relative to cPCA while providing interpretable, parts-based representations, and that GPU acceleration substantially reduces wall-clock time for iterative factorization methods without compromising clustering quality. In addition, as described in Supplementary Section ~\ref{sec:app_minibatch}, we leverage a minibatch training strategy to scale \model to much larger datasets. By updating the model via minibatch akin to the stochastic gradient descent in deep learning (Supplementary Section ~\ref{sec:app_minibatch}), we reduce memory demands and enable distributed or streaming analyses while maintaining stable clustering performance.


\section{Discussion}

Identifying interpretable latent structures enriched for target conditions relative to confounding backgrounds remains a central challenge in high-dimensional biological data analysis. The rapid expansion of high-throughput sequencing technologies has intensified the demand for methods that are both scalable and directly interpretable at the feature level. To address this need, we developed \model, which integrates a contrastive objective with the additive, parts-based representations of non-negative matrix factorization. By jointly modeling target and background datasets with a shared basis, \model isolates topics enriched in the target while attenuating variation expressed in both datasets. This formulation provides an interpretable alternative to variance-based contrastive approaches such as contrastive PCA, which emphasize differential variance but may lack feature-level interpretability, and to matrix factorization methods that yield interpretable topics but are not inherently contrastive or require explicit regression against known background covariates \cite{zhao2021learning}. Across simulations and multiple biological settings—including mouse proteomics, leukemia single-cell RNA sequencing, postmortem brain data, and cancer perturbation screens—\model recovered biologically meaningful patterns that were obscured or partially confounded under non-contrastive analyses. In some datasets, performance was comparable to strong contrastive baselines such as cPCA, whereas in others \model provided clearer separation or improved interpretability, indicating that its advantages depend on the structure of target-background relationships. Collectively, these results demonstrate that \model offers a scalable and interpretable framework for contrastive analysis across diverse modalities.

Despite these strengths, \model relies on several structural assumptions. The use of a shared basis matrix presumes that target and background datasets admit a common latent representation, and performance may degrade if this assumption is strongly violated. In addition, the quality and relevance of the background dataset are critical: if the background fails to capture shared variation, contrastive topics may reflect residual confounding rather than truly target-specific structure. When target and background distributions are highly similar, contrastive signal may be weak or difficult to disentangle. These considerations highlight that \model complements, rather than replaces, careful experimental design and background selection.

Beyond its current formulation, \model admits several methodological extensions. One direction is the integration of multi-modal or multi-view data, where modalities such as scRNA-seq and scATAC-seq, spatial transcriptomics and imaging, or transcriptomic and proteomic assays are analyzed jointly within a shared contrastive framework. Extending \model to such settings could enable extraction of shared and modality-specific contrastive topics, analogous to developments in multi-view contrastive learning \cite{chen2023deep, xu2023self}, while preserving the interpretability afforded by non-negative components. Another extension involves incorporating multiple background datasets or hierarchical contrastive designs to support structured comparisons across several biological conditions, for example healthy controls versus multiple disease subtypes. Such developments could build upon hierarchical and multi-reference contrastive analysis frameworks \cite{abid2018exploring, park2025systematic}. In addition, future work may explore incorporating explicit positive-pair contrast mechanisms within the factorization objective, drawing inspiration from recent extensions of contrastive principal component methods such as PCA++ \cite{wu2025pca++}, which leverage structured positive relationships to further enhance contrastive signal separation.

From a computational perspective, \model can in principle be adapted to online or streaming settings in which data are acquired sequentially and factorization must be updated incrementally. This is enabled by our minibatch training algorithm (Supplementary Section~\ref{sec:app_minibatch}) for \model and would be particularly relevant for large-scale initiatives such as cell atlas projects or longitudinal perturbation studies. Integration with deep generative models represents another promising avenue. Incorporating non-negative constraints within variational inference or autoencoder architectures may combine the interpretability of matrix factorization with the flexibility of deep models. Despite related efforts \cite{abid2019contrastive, weinberger2023isolating},  new avenues remain to be explored to broaden the scope of contrastive factorization in complex biological settings.

An important practical consideration is the selection of the contrastive weight parameter $\alpha$, which controls the balance between fitting target structure and suppressing background-associated variation. Analogous to a regularization parameter, $\alpha$ influences the granularity and specificity of the learned topics. In this study, $\alpha$ was selected using model-intrinsic criteria and stability analyses, and grid-based exploration was used to assess robustness across reasonable ranges. Nevertheless, principled and fully data-driven strategies for adaptive selection of $\alpha$ remain an important area for future work. A related methodological extension involves relaxing the shared-basis assumption and learning separate basis matrices for the target and background datasets ($W_\mathrm{target}$ and $W_\mathrm{background}$), coupled through alignment or regularization constraints. Such a formulation would permit explicit representation of shared and target-specific structure rather than inducing contrast solely through penalization, aligning with recent shared-private latent variable models \cite{martens2024disentangling, weinberger2023isolating}.

From a theoretical standpoint, \model is related to recent advances in contrastive dimension reduction. Formal frameworks have characterized the decomposition of target variation into shared and contrastive subspaces relative to a background, together with hypothesis tests and estimators for contrastive dimensionality \cite{hawke2024contrastive}. Subsequent work has proposed systematic background selection procedures based on subspace inclusion criteria \cite{park2025systematic}. While \model does not directly inherit the linear subspace guarantees of these methods, these theoretical developments provide a principled lens through which to interpret contrastive structure and motivate further theoretical analysis of non-negative contrastive factorizations.

\model is conceptually distinct from batch effect correction approaches widely used in genomics and single-cell analysis \cite{stuart2019comprehensive, haghverdi2018batch}. Batch correction seeks to remove unwanted technical or biological variation and align datasets into a harmonized representation. In contrast, \model explicitly leverages differences between datasets to highlight structure enriched in the target relative to a background. Rather than eliminating background-associated variation, it isolates topics that deviate from that background. In practice, batch correction and \model may be complementary: integration methods can reduce purely technical artifacts prior to contrastive analysis, whereas \model can subsequently isolate biologically meaningful deviations relative to an appropriate reference.

In summary, by embedding contrastive analysis within a non-negative factorization framework, \model provides an interpretable alternative to existing contrastive dimension reduction methods. Its ability to extract additive topics that remain directly linked to original features makes it particularly suitable for biological applications in which mechanistic interpretation is essential. Continued integration with theoretical advances in contrastive testing and background selection, together with empirical exploration in multi-modal and adaptive settings, may further establish \model as a flexible framework for disentangling shared and context-specific signals across high-dimensional data modalities.

\section{Methods}

\subsection{\model details}
We introduce background contrastive non-negative matrix factorization (\model), an extension of classical NMF that incorporates a background dataset to isolate latent structure specific to the target data. Given a non-negative target matrix $X \in \mathbb{R}_{\geq 0}^{M \times N_X}$ and a background matrix $Y \in \mathbb{R}_{\geq 0}^{M \times N_Y}$, \model learns a shared non-negative basis $W \in \mathbb{R}_{\geq 0}^{M \times K}$ and dataset-specific coefficients $H_X \in \mathbb{R}_{\geq 0}^{K \times N_X}, H_Y \in \mathbb{R}_{\geq 0}^{K \times N_Y}$. A contrast parameter $\alpha \geq 0$ controls the degree to which background structure is suppressed in the decomposition.

We minimize a contrastive loss of the form:
$$
\min _{W, H_X, H_Y \geq 0} J(W, H_X, H_Y) \equiv \mathcal{L}\left(X, W H_X\right)-\alpha \mathcal{L}\left(Y, W H_Y\right),
$$
where $\mathcal{L}(\cdot, \cdot)$ is a loss function or negative likelihood function appropriate to the data type. This general formulation allows adaptation to different likelihoods, including Gaussian (squared error), Poisson, and (zero-inflated) negative binomial models.

As a concrete example, we first consider the squared Frobenius norm:
$$
\mathcal{L}(A, B)=\|A-B\|_F^2=\operatorname{tr}\left[(A-B)^{\top}(A-B)\right],
$$
which leads to the expanded contrastive objective:
$$
\begin{aligned}
J = & \operatorname{tr}\left(X^{\top} X-\alpha Y^{\top} Y\right)-\operatorname{tr}\left(2 X^{\top} W H_X-2 \alpha Y^{\top} W H_Y\right) \\
& +\operatorname{tr}\left(H_X^{\top} W^{\top} W H_X-\alpha H_Y^{\top} W^{\top} W H_Y\right) .
\end{aligned}
$$

We optimize this objective using multiplicative updates derived from a general principle:
$$
\theta \leftarrow \theta \circ \frac{\nabla^{-}_{\theta} J}{\nabla^{+}_{\theta} J},
$$
where $\nabla^{-}_{\theta} J$ and $\nabla^{+}_{\theta} \mathcal{L}$ are the negative and positive parts of the gradient, and $\circ$ denotes element-wise multiplication. This update arises from a choice of adaptive learning rate,
$$
\eta_\theta=\frac{\theta}{\nabla_\theta^{+} J},
$$
When applied to a standard gradient descent step $\theta \leftarrow \theta-\eta_\theta \nabla_\theta J$, this yields the multiplicative update form. This scheme ensures non-negativity by construction and promotes convergence under mild conditions.

For the squared Frobenius loss, the gradients with respect to each topic matrix can be decomposed into positive and negative terms. Specifically, the gradient of the objective with respect to $H_X$ is:
$$
\nabla_{H_X} J=W^{\top} W H_X-W^{\top} X=\nabla^{+} H_X J-\nabla^{-} H_X J
$$
where $\nabla^{+} H_X J=W^{\top} W H_X$ and $\nabla^{-} H_X J=W^{\top} X$. Applying the multiplicative update rule yields:
$$
H_X \leftarrow H_X \circ \frac{W^{\top} X}{W^{\top} W H_X}
$$
Similarly, for $H_Y$, we have:
$$
H_Y \leftarrow H_Y \circ \frac{W^{\top} Y}{W^{\top} W H_Y},
$$
and for the shared basis matrix $W$, the gradient is:
$$
\nabla_W J=W H_X H_X^{\top}+\alpha Y H_Y^{\top}-X H_X^{\top}-\alpha W H_Y H_Y^{\top}
$$
with positive and negative parts given by:
$$
\nabla_W^{+} J=W H_X H_X^{\top}+\alpha Y H_Y^{\top}, \quad \nabla_W^{-} J=X H_X^{\top}+\alpha W H_Y H_Y^{\top}
$$
which gives the multiplicative update:
$$
W \leftarrow W \circ \frac{X H_X^{\top}+\alpha W H_Y H_Y^{\top}}{W H_X H_X^{\top}+\alpha Y H_Y^{\top}} .
$$

These updates are computationally efficient and retain non-negativity by construction. They serve as the foundation for extensions to other divergences (e.g., Poisson, Negative Binomial; Supplementary Section ~\ref{sec:app_distributions}) as well as mini-batch optimization schemes (Supplementary Section ~\ref{sec:app_minibatch}) that enable scalable training on large datasets. A formal convergence proof of the multiplicative-update algorithm is also provided in the Supplementary Materials (Supplementary Section~\ref{sec:app_convergence}).

\subsection{Selection of the contrastive parameter}

The contrast parameter $\alpha \geq 0$ controls the strength of background suppression in the contrastive objective, thereby governing the trade-off between fitting target-specific structure and penalizing variation shared with the background. Appropriate selection of $\alpha$ is therefore important for achieving effective contrast without over-suppressing informative signal in the target.

In our work, $\alpha$ is selected using criteria derived from the model fit itself, rather than downstream supervised or clustering-based performance, ensuring that tuning remains fully unsupervised and intrinsic to the contrastive factorization objective. Concretely, for a candidate value of $\alpha$, we assess target reconstruction together with background suppression by evaluating the contrastive objective at convergence. For the squared-error formulation, this corresponds to monitoring the contrastive reconstruction loss:
$$
J(\alpha)=|X-WH_X|_F^2-\alpha|Y-WH_Y|_F^2,
$$
and in likelihood-based formulations (e.g., Poisson, Negative Binomial, or Zero-Inflated Negative Binomial), the analogous contrastive negative log-likelihood is used. We exclude values of $\alpha$ for which the contrastive objective admits degenerate optima, such as collapse of the background coefficients $H_Y$ or severe loss of target reconstruction fidelity. This restriction ensures that $\alpha$ is selected within a regime where both target and background terms are simultaneously well-conditioned.

To explore the parameter space efficiently, we employ a coarse-to-fine line search over a predefined interval $[a,b]$, starting from small values of $\alpha$ and progressively increasing contrast strength. In practice, the objective typically stabilizes over a broad range of intermediate $\alpha$ values, whereas excessively large $\alpha$ leads to overly aggressive background suppression and diminished interpretability.

Across all experiments, we find that small to moderate values of $\alpha$ (typically $0.5$--$5$) yield stable factorizations that suppress shared variation while preserving target-enriched topics. Unless otherwise stated, we report results obtained using a representative value selected from this stable regime.

\subsection{Selection of the background dataset}

The effectiveness of \model critically depends on the selection of an appropriate background dataset. The goal is not to find a perfect negative control, but rather to identify a background that captures dominant sources of variation that are shared with, but not central to, the target dataset. By suppressing these shared topics, \model can more effectively isolate latent topics that are enriched in the target data.

The background may reflect biological, technical, or structural effects that confound interpretation. For example, when analyzing disease heterogeneity, a healthy control population can serve as a background to remove baseline inter-individual variation. In time-series data, early time points may act as a background to emphasize treatment or progression effects. Similarly, in tissue deconvolution, homogeneous cell populations may serve as background when identifying mixed-cell-type signals.

In practice, the background need not be drawn from the same experimental condition or matched one-to-one with the target. What matters is that the empirical structure of the background approximates the unwanted topics present in the target. In this sense, \model leverages the contrast between target and background distributions to guide factorization toward target-specific structure.

This idea is conceptually related to contrastive PCA, as described in \cite{abid2018exploring}, where the choice of background data follows a similar principle-capturing shared but uninformative variation to highlight structure enriched in the target dataset. However, \model extends this approach to the non-negative matrix factorization framework, enabling its application to a broader class of data types--including sparse and count-based data--while improving interpretability through parts-based representations.

In designing a background dataset, it is also important to consider representational balance. Overly noisy or under-sampled background data may not provide a reliable contrast, while an overly complex background may suppress informative signal. We therefore recommend empirical validation of the background’s effectiveness using held-out performance metrics or cross-condition reproducibility, in conjunction with interpretability of the learned topics.

\subsection{Data Preprocessing}

All datasets were preprocessed using standard procedures tailored to their data type. For single-cell RNA sequencing datasets, we began by applying quality control filters to remove low-quality cells and genes with negligible expression, following conventional best practices. Gene expression matrices were then restricted to protein-coding genes, and library-size normalization was performed to correct for sequencing depth differences across cells. Counts were subsequently log-transformed to stabilize variance, and the top 3,000 highly variable genes (HVGs) were selected using established variance-based heuristics to focus the analysis on the most informative features. For protein expression data, preprocessing involved the removal of proteins with low variance or excessive missingness, followed by normalization to correct for systematic differences in abundance across samples. All datasets were standardized so that target and background matrices were processed identically, ensuring that downstream contrastive analyses reflected biological or experimental differences rather than preprocessing artifacts.

\subsection{Comparison with Existing Methods}

To evaluate the performance of \model, we compared it against several widely used dimensionality reduction and contrastive analysis methods. Standard principal component analysis (PCA) was applied using the implementation provided in the scikit-learn Python package. Contrastive principal component analysis (cPCA) was performed using the official implementation released by the original authors in the contrastive Python package, which reproduces the algorithm described in \cite{abid2018exploring}.

For non-negative matrix factorization (NMF), we implemented the standard multiplicative update algorithm to minimize the Euclidean reconstruction error, following the classical formulation of Lee and Seung \cite{lee1999learning}. This ensured a consistent optimization scheme between NMF and \model, which also relies on multiplicative updates. All methods were applied to the same preprocessed data matrices, with the same number of topics and without additional downstream transformations, allowing for direct comparisons of their ability to recover target-specific structure.

\subsection{Statistical Testing}
Statistical tests were performed to assess the association between learned topics or gene/protein expression levels and experimental conditions of interest. For each topic obtained from \model, NMF, or cPCA, we compared topic usage between groups using two-sided two-sample t-tests. The resulting p-values were transformed into signed $-\log_{10} p$ values, with the sign determined by the direction of the mean difference between groups.

For gene and protein-level analyses, we applied the same two-sample t-test directly to the original expression matrix to evaluate whether high-loading genes or proteins identified by \model exhibited significant condition-specific differences in the raw data. No multiple testing correction was applied for these analyses, as they were intended for visualization and interpretation rather than formal differential expression testing.

To evaluate clustering performance, Adjusted Rand Index (ARI) scores were computed between unsupervised clusters derived from topic embeddings and the corresponding withheld ground-truth labels, providing a measure of alignment between inferred and true subgroup structure.

\subsection{Software Availability}

An open-source implementation of \model, including all algorithms, preprocessing pipelines, and analysis scripts used in this study, is available at \url{https://github.com/li-lab-mcgill/bcnmf}. The repository provides a Python package with detailed documentation, example notebooks reproducing the main analyses, and instructions for applying \model to new datasets. 

\subsection{Data Availability}

All datasets analyzed in this study are publicly available from their original sources. The mice protein expression dataset was obtained from previously published studies investigating Down syndrome-related molecular changes \cite{sarkar2008proteomic, geddes1990cases}. The single-cell RNA-seq leukemia dataset was generated by Zheng et al. \cite{zheng2017massively} and is available from 10x Genomics (patient 27, pre- and post-transplant bone marrow mononuclear cells). The major depressive disorder (MDD) single-nucleus RNA-seq dataset was published by Maitra et al. \cite{maitra2023cell} and profiles dorsolateral prefrontal cortex from well-characterized MDD patients and controls. The MIX-seq idasanutlin perturbation dataset was obtained from McFarland et al. \cite{mcfarland2020multiplexed}.

All data were downloaded directly from the original repositories or supplementary materials of the cited publications. No new datasets were generated in this work. Preprocessing was applied uniformly as described in the Methods. Detailed download instructions and preprocessing scripts for each dataset are provided in the \model code repository: \url{https://github.com/li-lab-mcgill/bcnmf}.

\section{Acknowledgment}
Y.L. is supported by Canada Research Chair (Tier 2) in Machine Learning for Genomics and Healthcare (CRC-2021-00547), Natural Sciences and Engineering Research Council (NSERC) Discovery Grant (RGPIN-2016-05174) and Canadian Institutes of Health Research (CIHR) Project Grant (PJT-540722). A.Y. is supported by NSERC Discovery Grant (RGPIN-2024-06780) and FRQNT Team Research Project Grant (FRQ-NT 327788).


\bibliographystyle{naturemag}

\bibliography{main}

\newpage
\appendix

\makeatletter
\renewcommand{\thesection}{S\@arabic\c@section}
\renewcommand{\thetable}{S\@arabic\c@table}
\renewcommand{\thefigure}{S\@arabic\c@figure}

\setcounter{section}{0}
\setcounter{figure}{0}
\setcounter{table}{0}

\section*{\centering Supplementary Materials}

\section{\model with other distributions}\label{sec:app_distributions}

\subsection{Poisson-based \model}\label{sec:app_poisson}
To model sparse count data such as single-cell RNA sequencing (scRNA-seq), we adopt a Poisson likelihood framework. Compared to squared-error loss, the Poisson likelihood more naturally captures the discrete and non-negative nature of count observations, yielding improved interpretability and robustness in biological settings.

Let $X \in \mathbb{R}_{\geq 0}^{M \times N_X}$ and $Y \in \mathbb{R}_{\geq 0}^{M \times N_Y}$ denote the non-negative count matrices for target and background samples, respectively, where $M$ is the number of features (e.g., genes) and $N_X, N_Y$ are the number of samples. We approximate these as low-rank products $W H_X$ and $W H_Y$, with shared basis $W \in \mathbb{R}_{\geq 0}^{M \times K}$ and sample-specific coefficient matrices $H_X \in \mathbb{R}_{\geq 0}^{K \times N_X}$ and $H_Y \in \mathbb{R}_{\geq 0}^{K \times N_Y}$.

Assuming element-wise Poisson observations, the (unnormalized) nagative log-likelihoods for $X$ and $Y$ under this factorization are given by:
$$\mathcal{L}_{\text{p}}\left(X, W H_X\right) \propto -X \circ \log \left(W H_X\right)+W H_X,$$
$$\mathcal{L}_{\text{p}}\left(Y, W H_Y\right) \propto -Y \circ \log \left(W H_Y\right)+W H_Y,$$

where all operations are element-wise.

We define a contrastive objective that maximizes the target likelihood (which equals to minimize the negative likelihood) while down-weighting the background signal:
$$J=\mathcal{L}_{\text{p}}\left(X, W H_X\right)-\alpha \mathcal{L}_{\text{p}}\left(Y, W H_Y\right)\propto - X \circ \log \left(W H_X\right)+ W H_X-\alpha\left(- Y \circ \log \left(W H_Y\right)+W H_Y\right),$$
where $\alpha>0$ is a tunable contrast parameter. To minimize this objective under non-negativity constraints, we derive multiplicative update rules for $H_X, H_{Y,}$ and $W$.

The update for the target-specific coefficients $H_X$ is
$$
H_X \leftarrow H_X \circ \frac{W^{\top}\left(\frac{X}{W H_X}\right)}{W^{\top} \mathbbm{1} },
$$
where $\mathbbm{1} \in \mathbb{R}^{M \times N_X}$ is a matrix of ones and division is element-wise. Similarly, the update for the background coefficients $H_Y$ is
$$
H_Y \leftarrow H_Y \circ \frac{W^{\top}\left(\frac{Y}{W H_Y}\right)}{W^{\top} \mathbbm{1} },
$$
with $\mathbbm{1} \in \mathbb{R}^{M \times N_Y}$. The update for the shared basis matrix $W$ combines both target and background terms:
$$
W \leftarrow W \circ \frac{\left(\frac{X}{W H_X}\right) H_X^{\top}+\alpha \mathbbm{1}  H_Y^{\top}}{\mathbbm{1}  H_X^{\top}+\alpha\left(\frac{Y}{W H_Y}\right) H_Y^{\top}},
$$
where the shapes of each term ensure compatibility: $\frac{X}{W H_X} \in \mathbb{R}^{M \times N_X}, H_X^{\top} \in \mathbb{R}^{N_X \times K}$, and thus the product lies in $\mathbb{R}^{M \times K}$, consistent with the shape of $W$.

These updates preserve non-negativity and ensure monotonic convergence under the Poisson model. By explicitly modeling count distributions, \model is particularly suited to scRNA-seq and other sequencing-based assays where direct modeling of discrete variability is needed.

\subsection{\model with Negative Binomial likelihood}\label{sec:app_nb}

For datasets characterized by overdispersed counts, such as single-cell RNA sequencing, the Negative Binomial (NB) likelihood offers a more flexible modeling framework than the Poisson distribution by explicitly accounting for variability beyond the mean.

Let $X \in \mathbb{R}_{\geq 0}^{M \times N_X}$ and $Y \in \mathbb{R}_{\geq 0}^{M \times N_Y}$ denote non-negative target and background count matrices, respectively. As in the Poisson model, we factorize these as $X \approx W H_X$ and $Y \approx W H_{Y,}$ where $W \in \mathbb{R}_{\geq 0}^{M \times K}$ is a shared basis matrix and $H_X, H_Y \in \mathbb{R}_{\geq 0}^{K \times N}$ are dataset-specific coefficient matrices. The scalar dispersion parameter $\theta>0$ is shared and assumed fixed.

Under the NB likelihood, the contrastive objective is defined as:
$$J = \mathcal{L}_{\text{NB}}(X, WH_X) - \alpha\, \mathcal{L}_{\text{NB}}(Y, WH_Y),$$
where $\alpha>0$ is the contrast weight. The unnormalized Negative Binomial's negative log-likelihood is proportion to:
$$\mathcal{L}_{\text{NB}}\left(X, W H_X\right) \propto - X \circ \log \left(W H_X\right)+(X+\theta) \circ \log \left(\theta+W H_X\right)$$
$$
\mathcal{L}_{\text{NB}}\left(Y, W H_Y\right) \propto -Y \circ \log \left(W H_Y\right)+(Y+\theta) \circ \log \left(\theta+W H_Y\right),
$$
where all operations are element-wise and $\theta>0$ is the (shared) dispersion parameter. To minimize $\mathcal{L}$, we derive the following multiplicative updates under non-negativity constraints.

The update for the target coefficients $H_X$ is
$$
H_X \leftarrow H_X \circ \frac{W^{\top}\left(\frac{X}{W H_X}\right)}{W^{\top}\left(\frac{\theta+X}{\theta+W H_X}\right)},
$$
and for the background coefficients $H_{Y_{,}}$
$$
H_Y \leftarrow H_Y \circ \frac{W^{\top}\left(\frac{Y}{W H_Y}\right)}{W^{\top}\left(\frac{\theta+Y}{\theta+W H_Y}\right)} .
$$
The gradient of $J$ with respect to $W$ is given by
$$
\frac{\partial J}{\partial W}=-\left(\frac{X}{W H_X}-\frac{\theta+X}{\theta+W H_X}\right) H_X^{\top}+\alpha\left(\frac{Y}{W H_Y}-\frac{\theta+Y}{\theta+W H_Y}\right) H_Y^{\top},
$$
yielding the multiplicative update for $W$:
$$
W \leftarrow W \circ \frac{\left(\frac{X}{W H_X}\right) H_X^{\top}+\alpha\left(\frac{\theta+Y}{\theta+W H_Y}\right) H_Y^{\top}}{\left(\frac{\theta+X}{\theta+W H_X}\right) H_X^{\top}+\alpha\left(\frac{Y}{W H_Y}\right) H_Y^{\top}} .
$$
These updates ensure non-negativity and stable convergence while explicitly accommodating overdispersed counts. The NB-based \model formulation is thus well-suited for high-variance biological data, where Poisson assumptions may underestimate true variability.

\subsection{\model with Zero-Inflated Negative Binomial likelihood}

In highly sparse count datasets such as single-cell RNA sequencing (scRNA-seq), zero inflation arises due to technical dropout and biological heterogeneity. The Zero-Inflated Negative Binomial (ZINB) distribution captures both overdispersion and the excess of zeros, providing a more expressive generative model than the standard NB.

Let $X \in \mathbb{R}_{\geq 0}^{M \times N_X}$ and $Y \in \mathbb{R}_{\geq 0}^{M \times N_Y}$ be the observed target and background count matrices, factorized as $W H_X$ and $W H_Y$ as before. The ZINB distribution introduces a dropout probability parameter $\pi \in[0,1]$, and the dispersion parameter $\theta>0$ remains fixed and shared across entries.

Let $I_X^{+}$ and $I_X^0$ denote binary indicator matrices where $\left[I_X^{+}\right]_{ij}=1$ if $X_{ij}>0$, and $\left[I_X^0\right]_{ij}=1$ if $X_{ij}=0$; similarly for $I_Y^{+}, I_Y^0$.

We first write the overall contrastive objective as:
$$J = \mathcal{L}_{\text{ZINB}}(X, WH_X) - \alpha\, \mathcal{L}_{\text{ZINB}}(Y, WH_Y),$$
The unnormalized negative loglikelihood is proportion to:
$$
\begin{aligned}
\mathcal{L}_{\text {ZINB }}\left(X, W H_X\right) \propto I_X^{+} \circ & \left(\log (1-\pi)+X \circ \log \left(W H_X\right)-(X+\theta) \circ \log \left(\theta+W H_X\right)\right) \\
& +I_X^0 \circ \log \left(\pi+(1-\pi) \circ\left(\frac{\theta}{\theta+W H_X}\right)^\theta\right)
\end{aligned}
$$
$$
\begin{gathered}
\mathcal{L}_{\text {ZINB }}\left(Y, W H_Y\right) \propto I_Y^{+} \circ\left(\log (1-\pi)+Y \circ \log \left(W H_Y\right)-(Y+\theta) \circ \log \left(\theta+W H_Y\right)\right) \\
+I_Y^0 \circ \log \left(\pi+(1-\pi) \circ\left(\frac{\theta}{\theta+W H_Y}\right)^\theta\right) .
\end{gathered}
$$
Use the update rule, we get the multiplicative update for the target coefficient matrix $H_X \in \mathbb{R}_{\geq 0}^{K \times N_X}$ is:
$$
H_X \leftarrow H_X \circ \frac{W^{\top}\left(I_X^{+} \circ \frac{X}{W H_X}\right)}{W^{\top}\left[I_X^{+} \circ \frac{\theta+X}{\theta+W H_X}+I_X^0 \circ \frac{(1-\pi) \theta\left(\frac{\theta}{\theta+W H_X}\right)^\theta}{\left(\theta+W H_X\right)\left[\pi+(1-\pi)\left(\frac{\theta}{\theta+W H_X}\right)^\theta\right]}\right]},
$$
and for the background coefficient matrix $H_Y \in \mathbb{R}_{\geq 0}^{K \times N_Y}$:
$$
H_Y \leftarrow H_Y \circ \frac{W^{\top}\left(I_Y^{+} \circ \frac{Y}{W H_Y}\right)}{W^{\top}\left[I_Y^{+} \circ \frac{\theta+Y}{\theta+W H_Y}+I_Y^0 \circ \frac{(1-\pi) \theta\left(\frac{\theta}{\theta+W H_Y}\right)^\theta}{\left(\theta+W H_Y\right)\left[\pi+(1-\pi)\left(\frac{\theta}{\theta+W H_Y}\right)^\theta\right]}\right]} .
$$

To get the update for the basis matrix $W \in \mathbb{R}_{\geq 0}^{M \times K}$, we calculate the gradient with respect to $W \in \mathbb{R}_{\geq 0}^{M \times K}$, which decomposes into positive and negative topics:
$$\frac{\partial J}{\partial W} = \left( \nabla \mathcal{L}_X^+ - \nabla \mathcal{L}_X^- \right) H_X^{\top} - \alpha \left( \nabla \mathcal{L}_Y^+ - \nabla \mathcal{L}_Y^- \right) H_Y^{\top}.$$
Following the multiplicative update strategy, we update W by element-wise division of negative over positive contributions:
$$W \leftarrow W \circ \frac{ \nabla \mathcal{L}_X^- H_X^{\top} + \alpha\, \nabla \mathcal{L}_Y^+ H_Y^{\top} }{ \nabla \mathcal{L}_X^+ H_X^{\top} + \alpha\, \nabla \mathcal{L}_Y^- H_Y^{\top} }.$$
Substituting the expressions explicitly, the update becomes:
$$W \leftarrow W \circ \frac{ \left[ I_X^+ \circ \frac{X}{WH_X} \right] H_X^\top + \alpha \left[ I_Y^+ \circ \frac{\theta + Y}{\theta + WH_Y} + I_Y^0 \circ \frac{(1 - \pi) \theta \left( \frac{\theta}{\theta + WH_Y} \right)^{\theta} }{ (\theta + WH_Y) \left[ \pi + (1 - \pi) \left( \frac{\theta}{\theta + WH_Y} \right)^{\theta} \right] } \right] H_Y^\top }{ \left[ I_X^+ \circ \frac{\theta + X}{\theta + WH_X} + I_X^0 \circ \frac{(1 - \pi) \theta \left( \frac{\theta}{\theta + WH_X} \right)^{\theta} }{ (\theta + WH_X) \left[ \pi + (1 - \pi) \left( \frac{\theta}{\theta + WH_X} \right)^{\theta} \right] } \right] H_X^\top + \alpha \left[ I_Y^+ \circ \frac{Y}{WH_Y} \right] H_Y^\top }.$$

This formulation explicitly separates signal (non-zero counts) from dropout events (zero counts), ensuring biologically informed and numerically stable updates. The ZINB-based \model formulation is particularly well-suited to ultra-sparse data settings, where both overdispersion and dropout must be accounted for simultaneously.

\section{Mini-batch optimization of contrastive NMF}\label{sec:app_minibatch}

For large-scale datasets, we adopt a mini-batch extension of contrastive non-negative matrix factorization (\model) under the Poisson likelihood. While standard multiplicative updates apply gradient steps to the full dataset, mini-batch processing enables efficient optimization and memory scaling. However, a key distinction arises: the shared basis matrix $W$ cannot be directly updated within each batch. Instead, contributions from all mini-batches must be accumulated across an iteration before performing a single update to $W$.

Let $X \in \mathbb{R}_{\geq 0}^{M \times N_X}$ and $Y \in \mathbb{R}_{\geq 0}^{M \times N_Y}$ denote the target and background data matrices, and let $K$ be the number of latent topics. Given a contrast weight $\alpha>0$, and a mini-batch size $B$, we factorize the data as $X \approx W H_X$ and $Y \approx W H_{Y,}$ with $W \in \mathbb{R}_{\geq 0}^{M \times K}, H_X \in \mathbb{R}_{\geq 0}^{K \times N_X}$, and $H_Y \in \mathbb{R}_{\geq 0}^{K \times N_Y}$. For each global iteration, the following steps are performed:

1. Coefficient Updates in Mini-batches. For each mini-batch of columns from $X$, denoted $X^{(b)} \in \mathbb{R}_{\geq 0}^{M \times B}$, the corresponding submatrix $H_X^{(b)} \in \mathbb{R}_{\geq 0}^{K \times B}$ is updated via the standard Poisson multiplicative rule:
$$
H_X^{(b)} \leftarrow H_X^{(b)} \circ \frac{W^{\top}\left(\frac{X^{(b)}}{W H_X^{(b)}}\right)}{W^{\top} \mathbbm{1}_{M \times B}},
$$
where division and multiplication are element-wise, and $1_{M \times B}$ denotes a matrix of ones.

2. Gradient Accumulation for $W$ (Target Term). Simultaneously, we accumulate the numerators and denominators needed for the eventual update of $W$:
$$
\texttt{W\_grad\_num} \mathrel{+}= \frac{X^{(b)}}{WH_X^{(b)}} (H_X^{(b)})^\top, $$
$$
\texttt{W\_grad\_den} \mathrel{+}= \mathbbm{1}_{M \times B} (H_X^{(b)})^\top.
$$

3. Repeat for Background Data. For each mini-batch of $Y^{(b)} \in \mathbb{R}_{\geq 0}^{M \times B}$, we update the corresponding coefficient matrix $H_Y^{(b)}$ as:
$$H_Y^{(b)} \leftarrow H_Y^{(b)} \circ \frac{W^{\top}\left(\frac{Y^{(b)}}{W H_Y^{(b)}}\right)}{W^{\top} \mathbbm{1}_{M \times B}},$$
and accumulate the background gradients as:
$$
\texttt{W\_grad\_num} \mathrel{+}= \alpha \cdot \mathbbm{\mathbbm{1}}_{M \times B} (H_Y^{(b)})^\top,
$$
$$
\texttt{W\_grad\_den} \mathrel{+}= \alpha \cdot \frac{Y^{(b)}}{WH_Y^{(b)}} (H_Y^{(b)})^\top.
$$

4. Single Update for $W$. After traversing all mini-batches, we update $W$ using the accumulated terms:
$$
W \leftarrow W \circ \frac{\texttt{W\_grad\_num}}{\texttt{W\_grad\_den}},
$$
where all operations are element-wise and the accumulated matrices have shape $M \times K$.

\section{Convergence proof of the multiplicative-update algorithm}\label{sec:app_convergence}

The multiplicative updates used in \model are derived to optimize a
contrastive objective under non-negativity constraints.
To ensure that the algorithm is well-founded, we prove that each update
monotonically decreases the objective function.
This guarantees stable optimization and validates the use of 
multiplicative rules in practice. We present a complete proof using the majorization function framework,
following the classical analysis \cite{lee2000algorithms},
extended to the contrastive setting.

\begin{definition}[Majorization function]
\label{def:auxiliary_function}
A function $G(Z, Z')$ is an majorization function for $F(Z)$ if
\begin{enumerate}
\item $G(Z, Z) \equiv F(Z)$ for all $Z$, and
\item $G(Z, Z') \ge F(Z)$ for all $Z, Z'$.
\end{enumerate}

Then, defining
$$
Z^{(t+1)} = \arg\min_{Z \ge 0} G\bigl(Z, Z^{(t)}\bigr),
$$
yields
$$
F\bigl(Z^{(t+1)}\bigr)
\le
G\bigl(Z^{(t+1)}, Z^{(t)}\bigr)
\le
G\bigl(Z^{(t)}, Z^{(t)}\bigr)
=
F\bigl(Z^{(t)}\bigr),
$$
so $F\bigl(Z^{(t)}\bigr)$ is monotonically non-increasing.
\end{definition}

This majorization function principle provides the foundation of our convergence analysis. Rather than minimizing the original objective $F$ directly, which may be difficult due to coupling between variables and non-negativity constraints, we instead construct a surrogate function $G$ that upper-bounds $F$ while touching it at the current iterate. By minimizing this simpler surrogate at each step, we guarantee a descent in the true objective. In the following, we apply this framework to each block variable in the contrastive NMF objective and explicitly construct suitable majorization functions that yield the multiplicative updates.

Fix $W$ and $H_Y$. The objective as a function of $H_X$ is
$$
J\left(H_X\right)=\left\|X-W H_X\right\|_F^2+\text { constant },
$$
since the background term does not depend on $H_X$. This is exactly the Euclidean NMF subproblem.

Let
$$
A:=W^{\top} W \in \mathbb{R}_{\geq 0}^{K \times K}, \quad B:=W^{\top} X \in \mathbb{R}_{\geq 0}^{K \times N_X} .
$$
Up to an additive constant independent of $H_{X},$
$$
J\left(H_X\right)=\operatorname{tr}\left(H_X^{\top} A H_X\right)-2 \operatorname{tr}\left(H_X^{\top} B\right) .
$$

\begin{lemma}[Auxiliary function for Euclidean NMF in $H$]
\label{lem:aux_euclidean_nmf_H}
For any current iterate $H_X^{(t)} > 0$, define
$$
G\bigl(H_X, H_X^{(t)}\bigr)
:=
J\bigl(H_X^{(t)}\bigr)
+
\sum_{k,n}
\frac{\bigl(A H_X^{(t)}\bigr)_{kn}}{H_{X,kn}^{(t)}}
\bigl(H_{X,kn} - H_{X,kn}^{(t)}\bigr)^2
+
\left\langle
\nabla J\bigl(H_X^{(t)}\bigr),
\, H_X - H_X^{(t)}
\right\rangle .
$$
Then $G\bigl(\cdot, H_X^{(t)}\bigr)$ is an majorization function for $J(\cdot)$ on the non-negative orthant.
\end{lemma}

This is the standard Lee-Seung Euclidean-NMF construction: the quadratic form $\operatorname{tr}\left(H^{\top} A H\right)$ is majorized by a separable weighted sum of squares using the inequality $$\sum_j A_{k j} h_j^2 \leq\left(\sum_j A_{k j} h_j^{(t)}\right) \sum_j \frac{h_j^2}{h_j^{(t)}},$$ applied entrywise and summed over columns.
Minimizing $G\left(H_X, H_X^{(t)}\right)$ w.r.t. each scalar $H_{X, k n} \geq 0$ yields the closed-form update
$$
H_X \leftarrow H_X \circ \frac{W^{\top} X}{W^{\top} W H_X},
$$
where all operations are element-wise. Therefore, this update satisfies 
$$J\left(W, H_X^{(t+1)}, H_Y^{(t)}\right) \leq J\left(W, H_X^{(t)}, H_Y^{(t)}\right).$$
The same argument applies to $H_Y$ with $X$ replaced by $Y$:
$$
H_Y \leftarrow H_Y \circ \frac{W^{\top} Y}{W^{\top} W H_Y} \quad \Longrightarrow \quad J\left(W, H_X^{(t+1)}, H_Y^{(t+1)}\right) \leq J\left(W, H_X^{(t+1)}, H_Y^{(t)}\right) .
$$

\clearpage
\section{Supplementary Figures}
\subsection{Extended enrichment analysis of MDD-associated topics}

\begin{figure}[h!]
\centering
\includegraphics[width=0.7\textwidth]{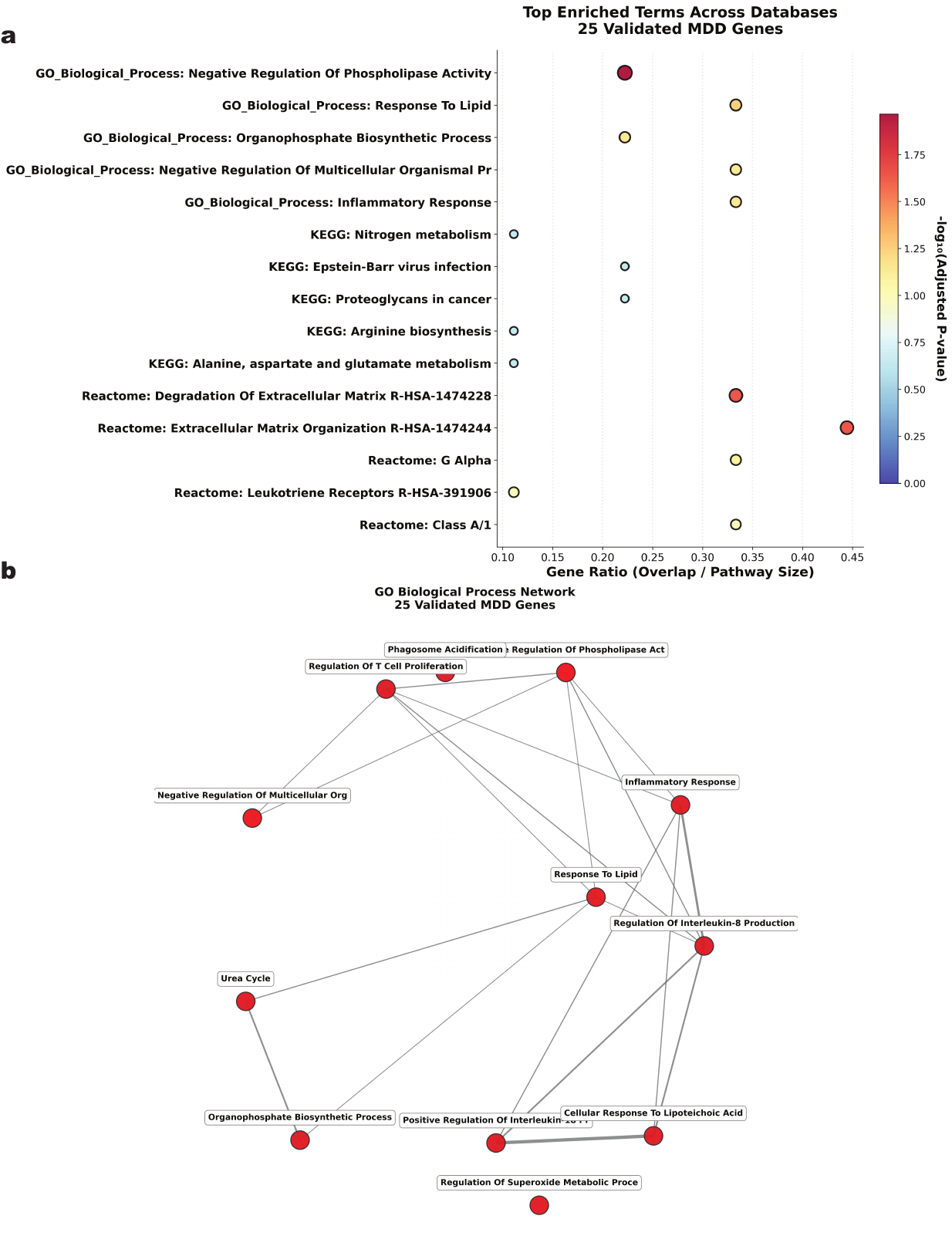}
\caption{
\textbf{Extended enrichment analysis of disease-associated gene programs in MDD.}
(a) Multi-database enrichment of 25 validated MDD-associated genes across GO Biological Process, KEGG, and Reactome. The x-axis shows gene ratio (overlap / pathway size); point size and color reflect statistical significance ($-\log_{10}$ adjusted $p$-value). Extracellular matrix and immune-inflammatory pathways show the strongest enrichment.
(b) Gene-overlap network of the top enriched GO Biological Process terms. Nodes represent enriched terms colored by significance; edges indicate Jaccard similarity $> 0.2$. A central immune-inflammatory module is observed, with metabolic terms positioned peripherally.
}

\label{fig:MDD_app}
\end{figure}

\end{document}